# Prior knowledge distillation based on financial time series

Jie Fang and Jianwu Lin

**Abstract**—One of the major characteristics of financial time series is that they contain a large amount of non-stationary noise, which is challenging for deep neural networks. People normally use various features to address this problem. However, the performance of these features depends on the choice of hyper-parameters. In this paper, we propose to use neural networks to represent these indicators and train a large network constructed of smaller networks as feature layers to fine-tune the prior knowledge represented by the indicators. During back propagation, prior knowledge is transferred from human logic to machine logic via gradient descent. Prior knowledge is the neural network's deep belief and teaches the network to not be affected by non-stationary noise. Moreover, co-distillation is applied to distill the structure into a much smaller size to reduce redundant features and the risk of overfitting. In addition, the decisions of the smaller networks in terms of gradient descent are more robust and cautious than those of large networks. In numerical experiments, we find that our algorithm is more accurate than traditional methods on real financial datasets. We also conduct experiments to verify and comprehend the experiment result.

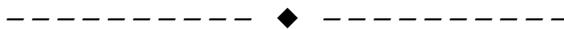

## 1 INTRODUCTION

One of the major characteristics of financial time series is that they contain a large amount of non-stationary noise [1]. In classification problems, deep neural networks (DNNs) are a very popular way to learn features from a data set and to weight each feature. However, some fake features may be abstracted from the data by mistake due to the high level of noise. Thus, when we model financial time series by means of DNNs, the noise may be treated as a feature, which is a major limitation of the models' generalization ability.

To address this problem, researchers often select quantitative indicators based on their financial experience, which is called prior knowledge (PK), as inputs for a neural network to construct features for the financial data. PK is usually obtained from experts in the given area and can make a model relatively concise and robust, which is called feature engineering. This type of methodology is common in the clinical and financial industries. [2][3] However, the performance depends substantially on the hyper-parameters chosen because the hyper-parameters cannot be adjusted in the training process. Therefore, an ideal method could represent this knowledge and allow the parameters to be tuned via training on actual data.

Different representations can help us entangle and hide various explanatory indicators of variation buried in data. Most scholars use probabilistic models and auto encoders to represent indicators to avoid a loss of variance [4]. Therefore, we use a neural network to represent PK. In addition, during backpropagation in the representation network, knowledge is transferred from one domain to the same domain because the joint distributions are the same but the logical meaning is slightly different. This structure is suitable for transfer learning because transfer learning is typically implemented when we want to use rich data in one domain to train a classifier in another domain. Most previously reported methods in transfer learning do not simultaneously reduce the difference in both the marginal distribution and conditional distribution between domains [5]. Feature-representation transfer learning does not suffer from the gap between the two distributions. Only the logical sense changes when we initialize the PK and after the backpropagation process.

However, many redundant features of financial time series coexist in certain indicators of human's PK about financial activities, which causes the DNNs constructed by feature-representation transfer learning based on redundant features cumbersome and computationally slow.

## 2 METHODOLOGY

We propose to use neural networks to represent financial features (prior knowledge). In this way, we will get a lot of sub networks, these sub networks can be merged into a large network, which contains all the knowledge. After combining these networks together, we will fine tune its hyper-parameters by minimizing the classification error. This large network is called the prior knowledge network (PKN), as shown in Figure 1a. The PK is the PKN's deep belief and teaches it not to be induced by the non-stationary noise.

However, the PKN suffers from another potential problem. Because we represent each feature as an individual network, the entire network structure containing all features is large. DNNs are commonly con-

- *Jie Fang is with the College of Information System, Tsinghua University, Shenzhen, China.*
  *E-mail: fangx18@mails.tsinghua.edu.cn.*
- *Jianwu Lin is with the College of Economic Management, Tsinghua University, Shenzhen, China.*
  *E-mail: lin.jianwu@sz.tsinghua.edu.cn.*

sidered to have excessive parameters, and these extra parameters lead to overfitting [6]. From our literature review, we find that distillation can efficiently reduce the number of parameters. To solve this problem, we implement the knowledge distillation method proposed by Hinton in 2015 to distill the entire structure into a much smaller one [7]. However, this approach does not provide satisfactory performance. However, the co-distillation proposed by Hinton in 2018 is effective [8]. We believe that the gradient descent of every epoch is important in this structure and represents how the teacher learns the PK and teaches students how to learn step by step. This kind of 'learn to learn' structure is a type of meta-learning. We propose to search for additional important findings in this area to improve the application of feature engineering. The entire learning process described above is called prior knowledge distillation (PKD), and the network generated by PKD is called the prior knowledge distillation network (PKDN), as shown in Figure 1b.

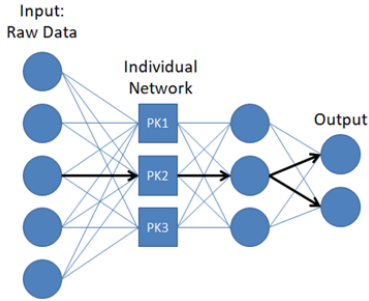

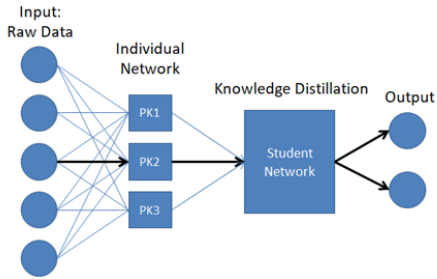

Fig. 1. Our Network structures: (a) PKN, (b) PKDN.

First, we pre-train each neural network with its learning target—PK represented by financial indicators, under the mean squared error loss function. Second, we combine the output of each neural network as new features. These outputs are all tensors and are not real numbers, which means that their values are tuned by backpropagation. Third, this feature layer goes through a fully connected layer to obtain a binary output, which can represent the probability of rising and falling. This is the teacher model, which we pre-train based on two distances. The first is the distance between the prediction and label, which is our target. The second is the distance between new PK learned by backpropagation and the initial PK. We let the teacher network struggle between these two distances, which can ensure stable gradient descent. In the next step, we train the small student networks individually. The last step is co-distillation. In each iteration, we obtain the output from the teacher network and the student network. For the loss function of the teacher model, we use the distance from the teacher and student and the distance between the new PK learned by the network and the initial PK. The loss function for students is set in the same way. The co-distillation process provides each network with the label information of another network, and it can also provide gradient descent information [8]. In this way, the knowledge from the teacher network can be passed to the student network; however, the simple structure of the student network reduces the likelihood of overfitting, which is the cornerstone of our algorithm's robustness.

## 3 NUMERICAL EXPERIMENTS

### 3.1 Real datasets

We use two real datasets to support our numerical experiments: the CSI500 Index and CSI300 Index, which are the most frequently traded indices in the A-share market. We use the past 50 minutes' index return series to perform binary classification of the next period of time. We perform tests for the next 1 minute, 3 minutes, 6 minutes, 9 minutes and 12 minutes.

For the experimental setting, we use 3000 samples as the training set, 300 samples as the validation set, and 300 samples as the test set. Because the financial time series is not stationary, all our experiments should be repeated for different periods of time. We repeat the experiment 10 times for the last 3 years, from 2016 to 2019, and use the mean and standard deviation to describe the performance, reported as the mean $\pm$ std, which can reduce the effect of randomness.

### 3.2 Representation of prior knowledge

Our numerical experiment uses two classic technical indicators in quantitative trading to serve as PK: simple moving average (SMA) and price rate of change (ROC). The formulas are shown in (1) (2). These two indicators use the momentum of returns of a financial index to perform prediction. [9] This PK is simple but very helpful for time-series classification, and it can also make our deep learning algorithms more explainable and make the decision process more transparent.

$$\text{SMA}(lag) = \frac{\sum_{i=n+1-lag}^{n} x_i}{lag} \quad (1)$$

$$\text{ROC}(lag) = \frac{x_n}{x_{n-lag}} - 1 \quad (2)$$

For each PK, lag is a decision variable, which we do not focus on here; we hope that the network can learn how long of a time series it needs and make this decision itself. Here, we only need to give the hyperparameter lag a reasonable initial value. As in the traditional quantitative trading method, we use a grid search to find the best hyper-parameters in the training set based on the highest information coefficient (IC). Normally, we use the Spearman coefficient of the indicator values and returns of financial instruments to represent IC, which can indicate how precisely we predict the return of financial instruments based on the indicator values.

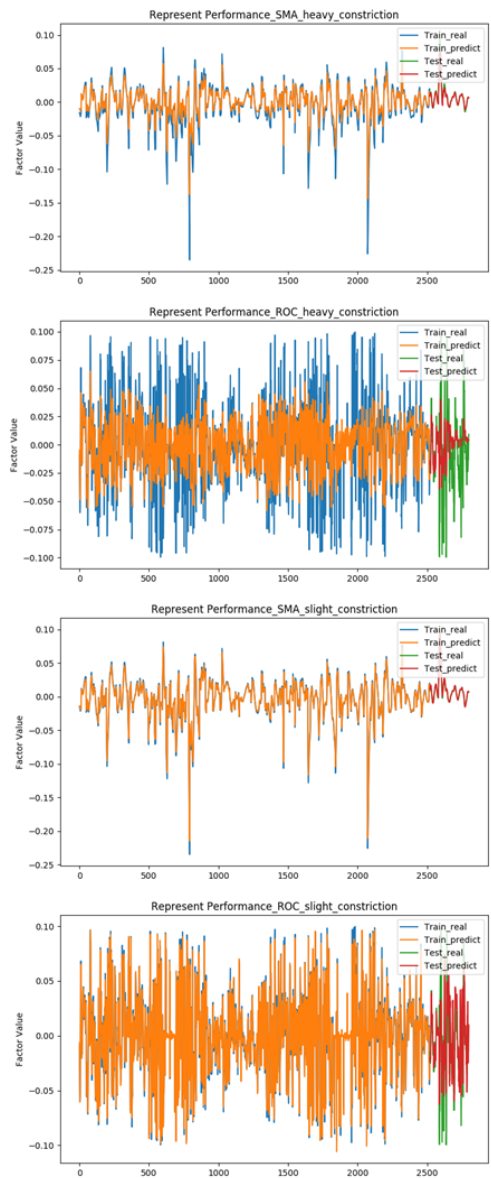

Fig. 2. The performance of representing prior knowledge with slight constriction and heavy constriction.

When we try to represent PK, we face a trade-off. If we use a large dropout rate and parameter penalization (heavy constriction) to learn the PK, we cannot represent the PK properly. By contrast, if the dropout rate and parameter penalization are very small (slight constriction), the PK is learned properly but overfitting occurs in the following learning process. Because the indicator value is very small, we can be easily misled by observing only the mean squared error. Thus, the representation performance by the neural network is plotted for both heavy constriction and slight constriction. These plots intuitively show the representation performance.

As shown in Figure 2, when the restriction is slight, the prediction and real value almost overlap. In summary, we successfully represent PK via a neural network under slight constriction. However, the PKN will be affected by overfitting due to its heavy network structure. We cannot rely on dropout and simple generalization methods to fix the overfitting problem, but no commonly used alternative techniques are available. Therefore, we must develop a tailored structure for this problem: we use co-distillation, as mentioned above.

### 3.3 Test results

The first advantage our PKDN is its fast testing speed due to its small network size. Faster testing speed is beneficial for mobile devices, such as finance apps. Moreover, high-frequency trading also has strict speed requirements. In our case, the smaller model has only 1/3 the number of parameters of the large model. We run our algorithms on Linux Sever with 64 GB RAM, 10G GPU 1080Ti and Tensorflow1.13. For PKN, it takes 325±53 microseconds for testing, but PKDN only takes 129±35 microseconds for testing.

The second strong point for PKDN is its relatively high test accuracy. We leverage the Spearman test in the training set to find suitable hyper-parameters for PK initialization. The experimental results are as follows.

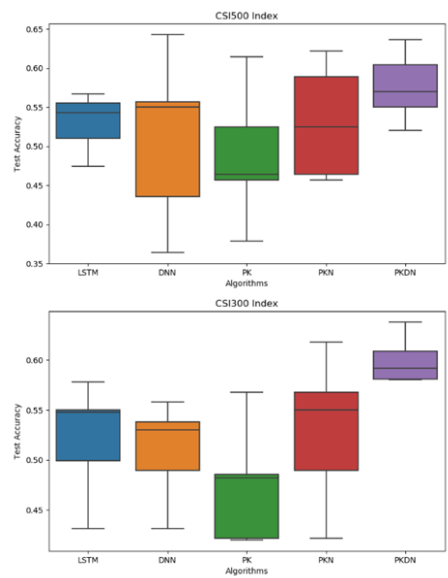

Fig. 3. Test accuracy on real financial datasets. We performed 10 individual experiments from January 2016 to May 2019.

We can draw three conclusions from Figure 3. First, DNN and LSTM [10] do not appear to provide good solutions to the financial time series problem. DNN performs poorly due to non-stationary noise, as does LSTM. LSTM is more powerful for addressing time series data: it is good at fitting the data but not good at identifying the tendency. Thus, we still need to change the content, loss function or structure to address financial time series.

Second, PKN sometimes performs better than PK but sometimes fails, which is why we say PKN has potential to outperform traditional methods. However, its training loss is much smaller than the testing loss. In the CSI300 Index dataset, this spread is not large; thus, the method performs well. Moreover, PKDN's training loss and testing loss are similar in almost all cases, and PKDN performs well on all these datasets.

Table 1. Test accuracy on real financial datasets (mean+-std). We performed 10 individual experiments from January 2016 to May 2019. We use the past 50 minutes of data to perform a binary classification for the next 1, 3, 6, 9 and 12 minutes.

| CSI300 | LSTM | DNN | PK | PKN | PKDN |
|---|---|---|---|---|---|
| Next 1 minutes | 0.580 ± 0.032 | 0.628 ± 0.037* | 0.495 ± 0.051 | 0.611 ± 0.022 | 0.587 ± 0.021 |
| Next 3 minutes | 0.541 ± 0.042 | 0.550 ± 0.082* | 0.494 ± 0.012 | 0.572 ± 0.016 | 0.533 ± 0.023 |
| Next 6 minutes | 0.521 ± 0.025 | 0.518 ± 0.031 | 0.487 ± 0.042 | 0.535 ± 0.041 | 0.585 ± 0.012* |
| Next 9 minutes | 0.466 ± 0.067 | 0.474 ± 0.064 | 0.519 ± 0.025 | 0.521 ± 0.055 | 0.564 ± 0.026* |
| Next 12 minutes | 0.486 ± 0.082 | 0.493 ± 0.027 | 0.515 ± 0.039 | 0.510 ± 0.025 | 0.544 ± 0.035* |

From Table 1, we find that DNN is powerful when we need to make a prediction for the next 1 or 3 minutes, which is trivial because the index will not change substantially within 1 or 3 minutes. The overfitting problem does not appear if we want to predict only in the short term. In addition, in the short term, our PK is not truly effective, as indicated by the performance of PK. However, when we make a long-term prediction, the index changes greatly enough to cover trading fees. In addition, the overfitting occurs in DNNs, but our PK works well: our PKN and PKDN methods show relative strength compared to the traditional method PK.

Since we do not know the true PK for the real-world time series, the test results may not fully illustrate the performance of PKD. To further support our idea and determine why it works, we conduct structured experiments.

## 4 STRUCTURED EXPERIMENTS

### 4.1 DNN has trouble learning financial time series

In this section, we use structured experiments to illustrate the advantages and disadvantages of each algorithm mentioned above. On the basis of these experiments, we explain why PKD outperforms the other algorithms. DNNs are powerful in many natural science fields, such as pattern recognition and speech recognition. In cases with a series composed of PK and unpredictable noise, a DNN (equipped with dropout, generalization, pruning and other basic techniques) will perform well if the noise is stationary. However, if the noise is non-stationary, the outstanding fitting ability of DNNs will become a burden. The training set includes a trap that negatively affects the DNN, but the real knowledge in the test sets is in the opposite direction. Here, we use two experiments to support our idea.

$$Y_{train} = Y_{test} = \text{SMA}(15,30) + \text{ROC}(15) + Noise * \text{Gaussian}(0,1) \quad (3)$$

$$\begin{cases} Y_{train} = \text{SMA}(15,30) + \text{ROC}(15) + Noise * (x(10) - x(15)) \\ Y_{test} = \text{SMA}(15,30) + \text{ROC}(15) + Noise * (x(15) - x(10)) \end{cases} \quad (4)$$

Let a Gaussian distribution with a mean of 0 and a standard deviation of 1 represent stationary noise, as shown in formula 3. Let the 10th price in the input series minus the 15th price in the input series present non-stationary noise. Noise is an independent variable that represents the noise level, as shown in formula 4. The label is determined by the value of Y: if Y is larger than 0, we set the label to 1; otherwise, we set the label to 0. As shown in Figure 4(a), the DNN outperforms all the other algorithms. When the Gaussian noise is large, the DNN still has high classification accuracy. This strong fitting ability, which we call 'machine logic', can handle stationary noise; however, as shown in Figure 4(b), the performance drops quickly. Because it has strong fitting ability, if we teach the DNN with non-stationary noise in the training set, the noise prevents it from learning the information in the time series completely. As a result, the DNN will fail in a testing set with an entirely different type of noise. In most cases, the noise in financial time series is non-stationary.

A DNN is good at fitting any formula, but it is difficult to prevent it from fitting the noise. With the help of PK and a reasonable network structure, we can let the neural network focus on PK but not fit the entire formula without restriction. As shown in Figure 4(b), the PK is simply our deep belief: as the noise level increases, the DNN fails completely. However, PKN and PKDN are highly stable. When the noise level is high, DNN will chase the non-stationary noise, but our algorithms focus on the deep belief.

### 4.2 Prior knowledge can be updated

The PK hyper-parameters can be optimized by training the PKN. Although we use the Spearman correlation to find a good initial hyper-parameter setting in training sets, the values can still be tuned by the powerful fully connected neural network. We conduct an experiment to illustrate this point.

$$Y_{train} = Y_{test} = \text{SMA}(15,30 + Noise) + \text{ROC}(15 + Noise) + \text{Gaussian}(0,1) \quad (5)$$

As shown in formula 5, we use 15 and 30 to initialize the SMA and 15 to initialize the ROC and add some noise to this parameter set. The results are shown in Figure 4(c): regardless of the hyper-parameter settings, PKDN always learns better than PK because it can fine-tune the distance, which is not possible in traditional methods. Sometimes, PKN performs worse than PK, likely because of the large network size.

### 4.3 Student networks can outperform teacher networks in some cases

Two networks are commonly used in knowledge distillation [8]: teacher networks and student networks. Normally, people use the teacher model to learn complicated tasks and let it teach much smaller student networks. In natural science fields, this technique is a popular way to accelerate a model. However, very few cases exist where the student networks outperform teacher network, which occurs in our case. The teacher network learns only PK and has a relatively large network size. The student network learns only the labels initially. The student networks have different knowledge, and we use co-distillation to let them learn from each other. The loss function is as follows:

$$\begin{cases} \text{Loss Teacher} = -\sum p_{real} * \log(p_{pred\ teacher}) - \sum p_{pred\ student} * \log(p_{pred\ teacher}) \\ \text{Loss Student} = -\sum p_{real} * \log(p_{pred\ student}) - \sum p_{pred\ teacher} * \log(p_{pred\ student}) \end{cases}$$

(6)

Because the student networks are small, we choose the student network as the final model. Furthermore, the student network has all the PK contained in the teacher network and is less likely to over fit the data, which is why the student network is so powerful. Finally, as mentioned in section 2, dropout and generalization will lower the representation performance based on PK. Thus, the overfitting problem is almost unavoidable in PKN, and distilling the teacher network into a smaller network is the tailored setting for this algorithm. To support this idea, let us consider Figure 4. In all cases, the test accuracy of PKDN is higher than that of PKN. As the noise level increases, the test accuracy of PKDN decreases more slowly than that of PKN, which illustrates the robustness of PKDN.

After analyzing these advantages, we consider a generalized situation that is common in the financial time series problem. In the real dataset experiment, we find that the Chinese stock index follows this structure. In this situation, our algorithm can show its strongpoints.

$$\begin{cases} Y_{train} = \text{SMA}(15,30) + \text{ROC}(15) + \text{Noise} * (x(10) - x(15)) + \text{Gaussian}(0,1) \\ Y_{test} = \text{SMA}(15,30) + \text{ROC}(15) + \text{Noise} * (x(15) - x(10)) + \text{Gaussian}(0,1) \end{cases}$$

(7)

DNN performs well initially but fails immediately when the non-stationary noise level increases. PK fluctuates greatly due to its fixed hyper-parameter settings. Sometimes, the PK adds up the noise to become a new distribution that is similar. In this case, PK performs well. However, this situation is just a coincidence. A more reasonable method is our PKDN, which learns prior knowledge and distills all the knowledge into a smaller network to overcome the overfitting problem in cases of both stationary and non-stationary noise.

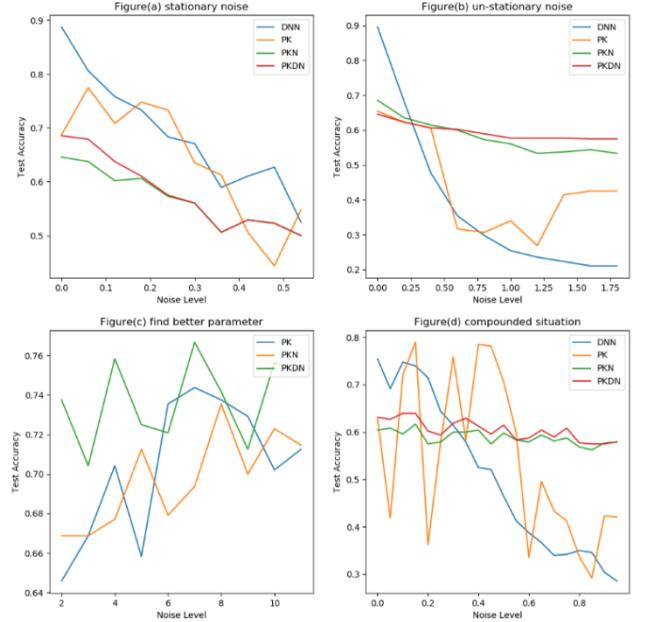

Fig. 4. Structured experiments to illustrate why DNN fails in non-stationary time series and why PKD and PKN are more powerful than PK.

## 4 CONCLUSIONS

In real quantitative trading, the DNN and LSTM are not as useful as our proposed method. DNN can over fit non-stationary noise, which makes it powerful in predicting short-term rises or falls but causes failure in the case of transaction fees. Moreover, research has shown that LSTM simply mimics the previous state in stock time series and cannot provide information about the tendency, which makes it almost useless in current quantitative trading algorithms.

Finance is a subject that has a high correlation with people's sentiment and intelligence. Moreover, finance is so unstable that powerful methods for pattern recognition and natural language processing perform poorly in this area. Our research is focused on this specific problem, and we develop a new algorithm that has both a neural network and the advantages of the traditional method. We think the prior knowledge can serve as the algorithm's deep belief. It will guide its gradient descent process and let it not be affected by the non-stationary noise. We believe that this method can provide a good basis for deep learning applications in quantitative trading, from the perspective of feature engineering.

**Jie Fang** is pursuing Master's Degree at Tsinghua Shenzhen International Graduate School. His research interests are financial feature engineering, reinforcement learning and meta learning.

**Jianwu Lin** received a dual Bachelor's degree in Engineering and Economics and a Master's of Engineering at Tsinghua University and a Ph.D. and dual master's degree at the University of Pennsylvania. Dr. Lin is a finance professor at Tsinghua Shenzhen International Graduate School. His research interests include Financial Technology, Financial Engineering, Meta Learning, Machine Learning, Behavioral Finance, Quantitative Investment and Robo-advising, Financial Risk Management, Internet Finance and Supply Chain Finance.